# Use of Computer Vision to Detect Tangles in Tangled Objects

Paritosh Parmar

Robotics Lab

SRM University

Chennai, India

paritosh_parmar@srmuniv.edu.in

*Abstract*—Untangling of structures like ropes and wires by autonomous robots can be useful in areas such as personal robotics, industries and electrical wiring & repairing by robots. This problem can be tackled by using computer vision system in robot. This paper proposes a computer vision based method for analyzing visual data acquired from camera for perceiving the overlap of wires, ropes, hoses i.e. detecting tangles. Information obtained after processing image according to the proposed method comprises of position of tangles in tangled object and which wire passes over which wire. This information can then be used to guide robot to untangle wire/s. Given an image, pre-processing is done to remove noise. Then edges of wire are detected. After that, the image is divided into smaller blocks and each block is checked for wire overlap/s and finding other relevant information. TANGLED-100 dataset was introduced, which consists of images of tangled linear deformable objects. Method discussed in here was tested on the TANGLED-100 dataset. Accuracy achieved during experiments was found to be 74.9%. Robotic simulations were carried out to demonstrate the use of the proposed method in applications of robot. Proposed method is a general method that can be used by robots working in different situations.

## I. INTRODUCTION

Automation and use of robots in personal space as well as industries is increasing every day. Some of the examples of personal robots are Roomba which can clean house on its own, robotic window cleaners, robotic lawn mower, and companion robots. A personal robot can be used to untangle structures like wires, threads, ropes and flexible pipes, hoses. Another usage of robots on similar line can be: electrical wiring and repairing on electric poles by robots. One of the main problems to be addressed in these cases is to detect tangles. This paper is the first to address detection of tangles in wires[1] using computer vision.

In a tangled mass of wires, wire/s pass over (or overlap) other wire/s or there may be one or more knots. Knots can also be taken as special case of overlap. Consider that wires are placed in X-Y plane. Consider for e.g. that in one tangle, wire A is passing over wire B. Therefore wire A is placed at a higher place along Z direction than wire B. But the difference in placements along Z direction is very small. Such a small difference can't be measured by currently available 3D cameras (which output RGB-D data). Therefore a method needs to be developed to detect tangles in a tangled mass of wires and also identify which wire is passing over the other wires using RGB images from 2-D camera. This information can then be used by robots to untangle the tangled mass.

In this paper, an algorithm to detect tangles is proposed. Given an image, following presented algorithm, tangles can be detected, and position (X, Y coordinates) of tangle can be found out. Also, which wire passes over other wire/s at a particular tangle can be found out using this algorithm. Deciding which wire passes over others requires an algorithm. An algorithm was developed for making this decision. This algorithm is important because the decision to be made is regarding the third coordinate axis i.e. Z-axis from available 2D data (RGB data from camera). The presented algorithm does not rely on or use any markers.

The method introduced in this paper was tested on 100 images. Test images had noise in them. Test images had "real world" background. Many images had multiple wires of different colors. Accuracy of 74.9% was achieved with the proposed method.

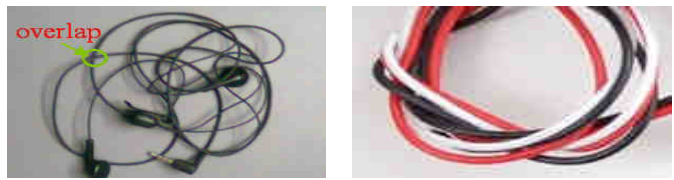

Fig. 1. Tangle/overlap indicated in a tangled mass of wire (left). Tangled mass of wires (right).

The remaining paper is organized in the following manner. Section II describes the overview of the methodology. Section III describes pre-processing. Section IV describes edge detection and thresholding. Then in Section V windowing approach and contour tracing are described. Post-processing and decision making process are presented in Section VI. Experiments and results are presented in Section VII. Section VIII concludes the paper.

## II. OVERVIEW OF THE METHODOLOGY

Images of tangled wires are acquired from a camera. In real world scenario, more than one wire can be tangled together. These wires can be of different colors. Out of these wires only one or some wires may be of interest. In that case, wire of interest is separated from others by using color separation technique. Separation of wire of interest is explained in detail in Section III. After separating out wire of interest, image is pre-processed for removing noise (smoothing). Removal of noise is important for edge detection and thresholding, to be carried out in subsequent steps. Removal of noise is explained in detail in Section III.

---

[1]Note: In this paper, words wire, pipe/hose, ropes, threads would equivalently be used for tangled objects.

After pre-processing images, edge detection is carried out on the image as a result of which boundaries of wire/s are detected. Edge detection is done by convolving the image with the kernel. In this method the image is convolved with 8 different kernels, one by one. Following steps are repeated for each kernel. The resultant image is then thresholded to obtain a binary image in which the edges are above the threshold value. In this way, the edges are separated out. Edge detection process and thresholding are explained in detail in Section IV.

After thresholding, 'sliding window' approach is taken. As a result of sliding window approach, small patches of edges are 'captured' in the sliding window. More than one patch of edges can be captured into the window. If only one patch is captured then there is no tangle in the area covered by the window. If two or more patches are obtained then there is a chance of a tangle in the nearby area. However, decisions regarding tangle are taken in subsequent steps. Following steps are repeated for every window. After patches of edge/s are captured into the window, contours are traced around those patches. Section V provides the details regarding window approach and contour tracing.

Contours developed are a sequence of *n - 1* points around the patches. Now *first* & *(n-2)*$^{th}$ point, *second* & *(n-3)*$^{th}$ i.e. $i^{th}$ and *(n-2-i)*$^{th}$ points are considered as pairs. Mid-point of two elements of each pair is calculated. Mean of these mid-points is calculated. A polynomial of degree *m*, fitting mid-points previously calculated, is formed. This process is carried out for every patch. Now, if more than one patch is present in a window, point of intersection of polynomials associated with these patches is calculated. Point of intersection gives the position of tangle i.e. coordinates at which two or more wires overlap which is nothing but a tangle. Distance of mean of mid-points from point of intersection, *d*, is calculated. Distance, *d*, is calculated for every patch present in a window.

Now, which wire passes over other wire/s is decided. Section VI explains in detail about contour development, and decision making process.

## III. PRE-PROCESSING

In some cases, multiple wires each with different color may be present. Of these multiple wires only one or some wires may be of interest. In such cases, further processing should be focused on wire/s of interest. For this, wire of interest is differentiated from the other wires based on its color. Images are acquired from a RGB digital camera. Since RGB is an additive color model, a wide range of colors are reproduced by superimposition (addition) of primary (red, blue, green) colors. An RGB image having width $w_I$ and height $h_I$ is $w_I$ x $h_I$ x *3* array which stores blue, red, and green components for each and every pixel. Color of a pixel is the result of the superimposition of green, red and blue colors present at that pixel position. Pure red color has red intensity of 255, pure green color has green intensity of 255 and likewise pure blue color has blue intensity of 255. For e.g. brown color has red color intensity of 185, green color intensity of 122, and blue color intensity of 87. Therefore for differentiating wire of the desired color, each pixel is checked for desired red, green, and blue color intensities. Value of only the matching pixels is set more than that of the black color. Hence, desired color, and consequently wire is differentiated. Fig. 2 shows a case in which multiple wire of different colors were present, of which only blue colored wire was of interest. Blue colored was differentiated from other wires.

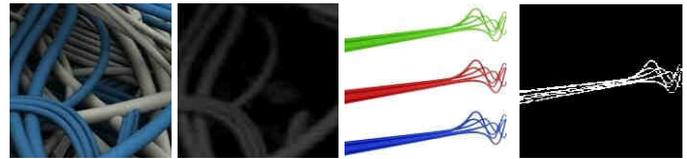

Fig. 2. Multiple wires of different colors (left image). Blue colored wire differentiated from rest of the wires (second image). Red colored wire differentiated (right image) from three wires (third image).

Once the color filtering is done, the image is smoothened to remove noise from the image. In the later stages, edge detection will be carried out. It is therefore, very necessary to remove noise from the image. During edge detection, derivatives of images will be taken. Any noise present in the image will get amplified when its derivative is taken [1]. To get rid of noise amplification problem, image is blurred [1], using Gaussian blur. In Gaussian smoothing, transformation to be applied to each and every pixel is calculated using two-dimensional Gaussian function. If *x* is the horizontal distance with respect to origin, and *y* is the vertical distance with respect to origin, then two-dimensional Gaussian function is defined as:

$$G(x,y) = \frac{1}{2\pi\sigma^2} e^{-\frac{x^2+y^2}{2\sigma^2}} \qquad (1)$$

$\sigma^2$ is variance [2] [3]. A kernel is built from the values obtained from Gaussian distribution. Amount of blurring produced is proportional to the dimensions of the kernel. Blurred image is obtained by convolving image with the kernel. New value of a pixel is the weighted average of that pixel and its neighbors. Highest weightage is given to the pixel under consideration. Weightage given to the neighboring pixels decreases as their distance from the pixel under consideration increases. Compared to uniform smoothing, Gaussian smoothing is more suitable for the proposed method because it preserves edges in a more effective manner. Preservation of the boundaries and the edges is essential for subsequent steps.

Image blurring is more useful when tangled object has very rough texture or is made by plying (twisting together multiple strands). Fig. 3 shows a rope which is made by twisting together multiple single strands.

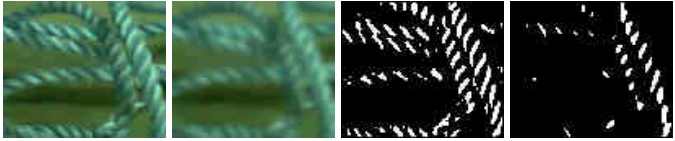

Fig. 3. First image is the original image. Second is the result blurring on the first image. Third and fourth images show edges detected in first and second image respectively.

From the case showed in Fig. 3, it is clear that blurring decreases the number of edges detected. When blurring is not applied in case of rough objects, unnecessary & increased number of edges would be detected.

## IV. EDGE DETECTION AND THRESHOLDING

After the pre-processing is completed, edge detection is carried out on the image. Edge detection is accomplished by convolving the pre-processed image with the eight Robinson compass masks. Robinson compass masks have 0, 1, and 2 as their coefficients [4]. Eight Robinson compass masks are obtained by taking a single a mask and rotating it to eight different directions, namely South (S), West (W), North (N), East (E), South-West (SW), North-East (NE), South-East (SE), and North-West (NW) [5] [6]. Table 1 shows eight Robinson Compass Masks.

TABLE 1: 8 ROBINSON COMPASS MASKS

| $\begin{bmatrix} -1 & -2 & -1 \\ 0 & 0 & 0 \\ 1 & 2 & 1 \end{bmatrix}$ (N) | $\begin{bmatrix} 1 & 2 & 1 \\ 0 & 0 & 0 \\ -1 & -2 & -1 \end{bmatrix}$ (S) |
|---|---|
| $\begin{bmatrix} 1 & 0 & -1 \\ 2 & 0 & -2 \\ 1 & 1 & -2 \end{bmatrix}$ (E) | $\begin{bmatrix} -1 & 0 & 1 \\ -2 & 0 & 2 \\ -1 & 0 & 1 \end{bmatrix}$ (W) |
| $\begin{bmatrix} 0 & -1 & -2 \\ 1 & 0 & -1 \\ 2 & 1 & 0 \end{bmatrix}$ (NE) | $\begin{bmatrix} -2 & -1 & 0 \\ -1 & 0 & 1 \\ 0 & 1 & 2 \end{bmatrix}$ (NW) |
| $\begin{bmatrix} 2 & 1 & 0 \\ 1 & 0 & -1 \\ 0 & -1 & -2 \end{bmatrix}$ (SE) | $\begin{bmatrix} 0 & 1 & 2 \\ -1 & 0 & 1 \\ -2 & -1 & 0 \end{bmatrix}$ (SW) |

Each mask is sensitive to edges oriented along only one of the eight directions [7]. The wire/s forming tangles would be oriented along different directions. Since each mask is sensitive to one direction, it helps in developing edge for certain segments of wire with each mask. Since each one of the eight masks is convolved with the given image, in all 8 resultant images would be obtained. Following procedure is carried out for each of the 8 resultant images.

After edge detection, the image is thresholded to a threshold level. Thresholding is done according to the method introduced by Otsu, in which, histogram shape-based image thresholding is automatically performed [8]. The algorithm assumes that the image to be thresholded contains two classes of pixels or bi-modal histogram. Two classes of pixels represent background and foreground. The algorithm calculates the threshold which optimally separates two classes of pixels such that their combined spread is minimized [9]. *Within-class* variance is defined as:

$$\sigma_{within}(T) = n_B(T)\sigma_B^2(T) + n_O(T)\sigma_O^2(T) \quad (2)$$

where,

$$n_B(T) = \sum_{i=0}^{T-1} p(i) \quad (3)$$

$$n_O(T) = \sum_{i=T}^{N-1} p(i) \quad (4)$$

$T$ is threshold level. $\sigma_B^2(T)$ is variance of pixels which are above threshold, $\sigma_B^2(T)$ is the variance of the pixels which are below threshold. Intensity levels range from 0 to *N-1*. *Between-class* variance is defined as:

$$\sigma_{between}^2(T) = \sigma^2 - \sigma_{within}^2(T) \quad (5)$$

where, $\sigma^2$ is the combined variance.

For each potential threshold T, following steps are carried out:

1) Pixels are separated into two groups.
2) Mean of each group is found out.
3) Difference between the means is found out.
4) Number of pixels in one group is multiplied by the number of pixels in the other group.

At optimum threshold level, *between-class* variance is maximized.

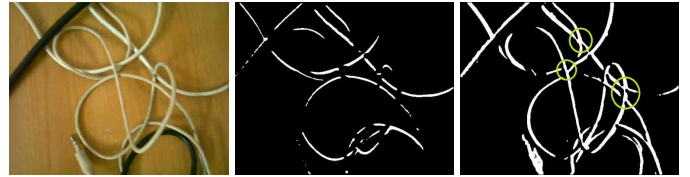

Fig. 4. First image shows original image. Second image shows the result of edge detection using N compass and thresholding performed on first image. Third image shows result of edge detection using E compass mask and thresholding performed on first image. Differences in the edges detected are clearly visible. Some of the wire overlapping (tangles) occurring in third image but not in second image are clearly marked by green colored circles.

## V. WINDOWING AND CONTOUR TRACING

### A. Windowing

Now, that the edges are detected and separated from their background, we will move on to detect tangles. For detecting tangles, we will search for two or more patches of the thresholded edges and determine if they intersect (and equally, overlap) when they are extrapolated. Some of the e.g. of patches and probable overlaps are shown in Fig. 5. Since analyzing the whole image at one go will not be fruitful therefore windowing approach is taken.

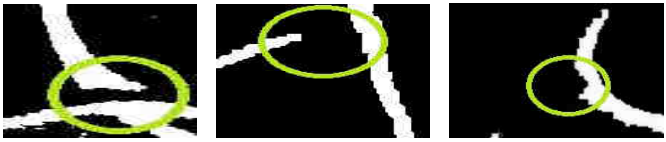

Fig. 5. Images show patches of detected & thresholded edges. Gaps marked by circles show position of probable overlap of wires i.e. a probable tangle.

Windowing approach can be explained as follows. A window of width, w and height, h is considered. Length of the diagonal is $d_w$. Starting from the top-left corner of image, the window is slide in right direction in a manner similar to raster scan.

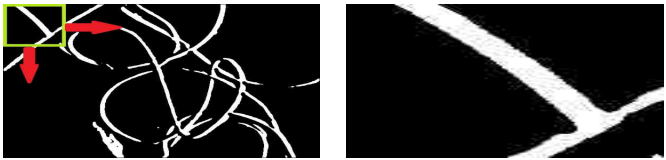

Fig. 6. Window and its sliding are shown (left image). Part of the image captured in the window (right image).

The image is subdivided into a sequence of horizontal strips known as 'scan rows'. The window sweeps/slides horizontally left-to-right, then stops when window reaches horizontal limit of the image and rapidly moves back to the left, where it turns back on and sweeps out the next row. While the window slides, it 'captures' a small part of the whole image. Parts of image 'captured' in window are processed further. Following steps (i.e. contour formation and post-processing (Section VI)) are performed for each part of the image captured by the window.

### B. Contour Formation

For each of the patches 'captured' in a window, contours are traced around them. A contour is a list of points that define a curve in an image [10]. Following algorithm was used for tracing contours around patches:

1) Start to search the image from top-left pixel until a pixel, $P_{C0}$ belonging to a patch (new region) is encountered. To store the direction of the previous move along the contour from previous move along the contour to the current contour element, a variable, *direction* is defined. *direction* is assigned following values:
    a) If the contour is detected in four-connectivity, *direction* = 0. Refer Fig. 7(a).
    b) If the contour is detected in eight-connectivity, *direction* = 7. Refer Fig. 7(b).
2) The 3 by 3 neighborhood of the pixel under consideration is searched in counter-clockwise direction. Start the search at the pixel positioned in the direction:
    a) (*direction* + 3) *modulo* 4. Refer Fig. 7(c).
    b) (*direction* + 6) *modulo* 8 if *direction* is odd. Refer Fig. 7(e). (*direction* + 7) *modulo* 8 if *direction* is even. Refer Fig. 7(d).
    The first pixel whose value is equal to current pixel is a contour element; $P_{Cn}$. Value of *direction* is updated.
3) Stop if $P_{Cn} = P_{C1}$ and $P_{C(n-1)} = P_{C0}$, else go to step *2*. Pixels starting from $P_{C0}$ to $P_{C(n-2)}$ constitute a contour.

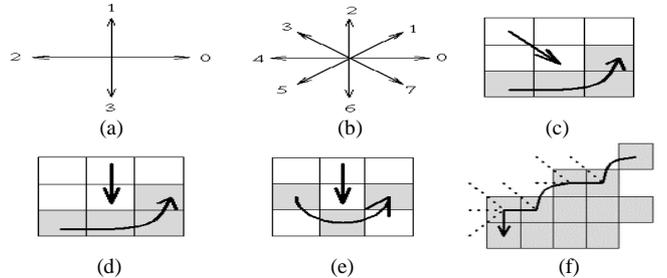

Fig. 7. Four-connectivity direction notification (a), eight-connectivity direction notification (b), pixel neighborhood search sequence in eight-connectivity (c), pixel neighborhood search sequence in eight-connectivity (d), pixel search sequence in four-connectivity (e), contour tracing in eight-connectivity (dashed lines show pixels tested during the contour tracing) (f).

### VI. POST-PROCESSING AND DECISION MAKING

At this stage, we have contour points, starting from $P_0$ to $P_{n-2}$, for each of the contour present in the window. Now, following steps are carried out for each of the patches:

1) $i^{th}$ and $(n-2-i)^{th}$ points are paired together. $i$ ranges from 0 to $\frac{n-2}{2}$. In all, $(\frac{n-2}{2} + 1)$ pairs are obtained.
2) Mid-points of pairs obtained in step *1* are calculated. Hence, $\frac{n-2}{2} + 1$ mid-points will be obtained. Let the mid-point of $k^{th}$ pair be represented by $(x_m^k, y_m^k)$.
3) Mean of the points obtained in step *2* is calculated. Let mean be represented by $X_m = \frac{\sum_{k=0}^{\frac{n-2}{2}+1} x_m^k}{\frac{n-2}{2}+1}$, $Y_m = \frac{\sum_{k=0}^{\frac{n-2}{2}+1} y_m^k}{\frac{n-2}{2}+1}$.
4) A polynomial of degree *m* is fitted to the points obtained in step *2* using Method of Least Squares. Following algorithm is used for determining optimum value of m:
    a) Take m = 5.
    b) If the polynomial of degree m fits the data, go to step *d*, else go to step *c*.
    c) If m = 1 go to step *d*, else decrease m by 1 and go to step *b*.
    d) End.

This algorithm is used to avoid *polynomial wiggle* [11].

Now, point of intersection of the polynomials found in *4* is calculated. Since patches are associated the wires, point of intersection safely provides position of the tangle. Distance, $d_j$, of mean of mid-points of $j^{th}$ patch, from point of intersection is calculated. Now, which wire passes over other wire/s at the position of tangle is to be decided. Patch corresponding to the wire overlapping other wires is nearest to the point of intersection (refer Fig. 5). Therefore patch whose corresponding $d_j$ is least among all the patches present in a window is taken (decided) to be patch belonging wire which overlaps the other wire/s. A confidence level is associated with every decision made. Confidence level is equal to the ratio of $d_w - d_j$ to $d_w$ (refer Section V. *A* for $d_w$). Remember the whole process is repeated 8 times, each time with different kernel (starting from convolution with different kernel for edge detection). So, there will be at most 8 decisions (assuming a tangle is detected in every iteration), each with a confidence level, for a tangle. Decision with the highest confidence level is chosen as the final decision.

## VII.  EXPERIMENTS AND RESULTS

### A. Data

A dataset named TANGLED-100, consisting of images tangled linear objects was introduced. The method presented in this paper was tested on TANGLED-100 dataset. Images in the dataset were acquired from a digital RGB camera. The dataset can be subdivided into two classes on the basis of the size of images: first class contained 70 images of size 640 x 480 pixels and the other class contained 30 images of size 1280x720 pixels. Images in the dataset have the following variations among themselves:

- Images were acquired in different environments; indoor and outdoor therefore have different levels of lighting and different background.
- Images contain different tangled objects such as wires, ropes, pipes/hoses. Thickness & length of these objects also varied.
- Multiple objects of different colors are present in images.

The dataset will be available from *http://www.computervisiononline.com/datasets* for research purposes.

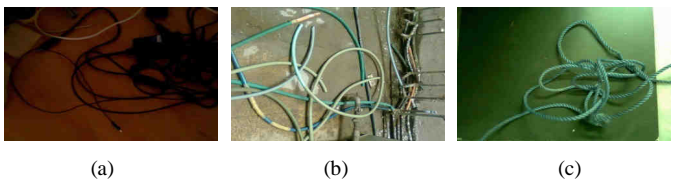

Fig. 8. Indoor image of a tangled wire (a), outdoor image of tangled pipes/hoses (b), indoor image of a tangled rope (c).

### B. Tools

MATLAB 7.10.0.499, OpenCV 2.4.2 library, Visual Studio 2010, and ABB RobotStudio 5.15 were used in carrying out experiments.

### C. Edge Detection and Thresholding Results

Images were pre-processed as described in Section III to remove noise. Level of blurring used for outdoor images is higher than that used for indoor images.

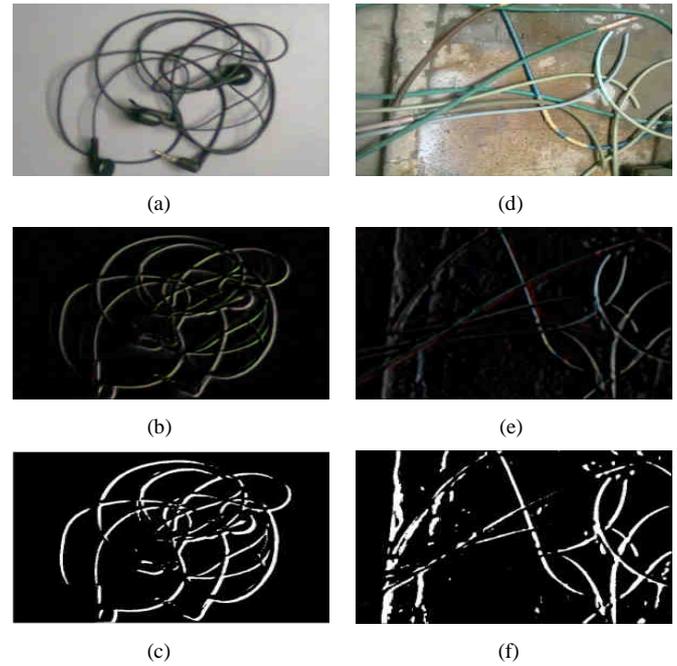

Fig. 9. (a) (b) (c) are indoor images. (d) (e) (f) are outdoor images. (a) (d) pre-processed images, (b) (e) edges detected, (c) (f) images (b) (e) are thresholded.

### D. Windowing and Contour Formation Results

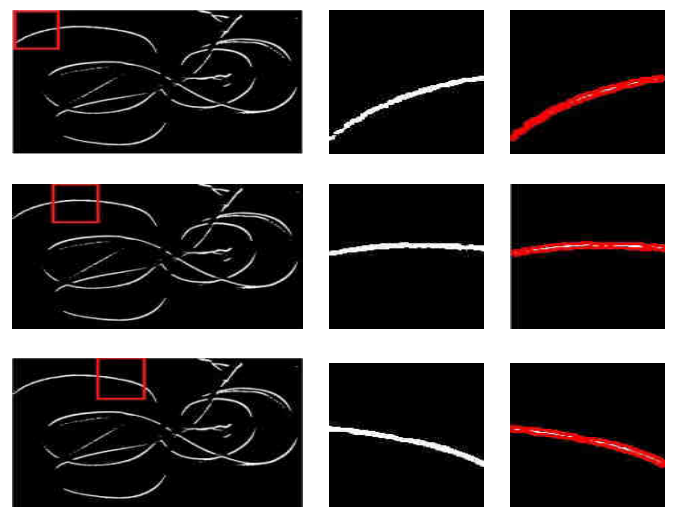

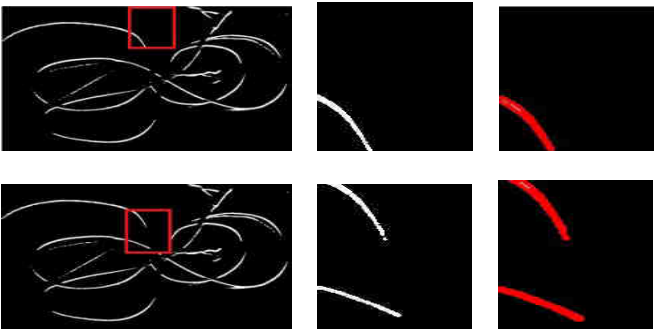

Fig. 10. Window sliding over the image (first column), patches captured by sliding window (second column), and contours traced around patches (marked in red color) (third column). Since two patches are captured in the last row, it indicates a possibility of tangle in the nearby area, which indeed is the case.

*E. Post-processing and Decision Making Results*

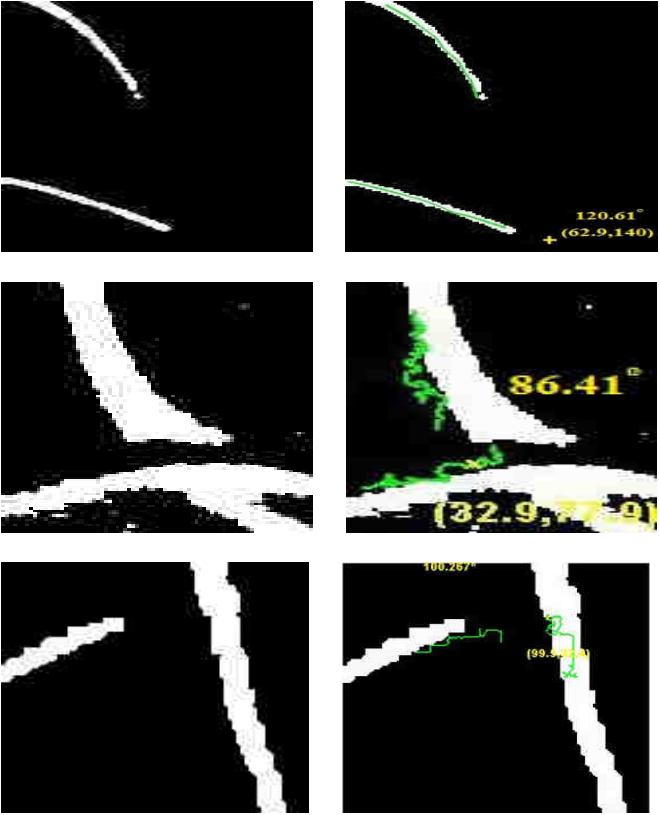

Fig. 11. Patches captured in window (first column), Polynomials fitting mid-points of contours are shown in green and point of intersection is shown by yellow cross hair (second column). Position of tangle is mentioned in brackets. Additional information regarding angle between patches at the point of intersection is mentioned.

Overall accuracy was measured as:

$$Accuracy = \frac{TP + TN}{TP + TN + FP + FN} \quad (6)$$

TABLE 2: ACCURACY OF ALGORITHM

| TP | TN | FP | FN | Accuracy |
|---|---|---|---|---|
| 0.479 | 0.270 | 0.175 | 0.075 | 0.749 |

*F. Robotic Application Simulation*

Fifteen basic robotic simulations were carried out using ABB RobotStudio 5.15. In these simulations, arm of IRB1600 Type A robot was made to travel to the position of the tangle.

VIII. CONCLUSION

In this paper, a method for detecting tangles, determining position of the tangle, and which wire passes over other wire at the tangle was proposed. The tangled objects we considered were wires, ropes, pipes/hoses, threads. The proposed method was tested on TANGLED-100 dataset, which was introduced with this method. The data set contained images of different objects having different properties and placed in different environments. Accuracy of 74.9% was achieved with this method. Finally, robot simulations were carried out to demonstrate the usage of this method in the field of robotics. In this paper, it is shown that it is important to use Robinson masks for edge detection and take a windowing approach. The information which is output from this method can be used by personal robots such as the PR2 by Willow Garage for untangling tangled objects. This can be very useful in assisting people as well as in industrial application. However, parallelizing this method can make this method faster.